\documentclass{article}

\usepackage{microtype}
\usepackage{graphicx}
\usepackage{subfigure}
\usepackage{booktabs} 

\PassOptionsToPackage{hyphens}{url}\usepackage{hyperref}

\usepackage[accepted]{templates/icml/icml2025}

\usepackage{amsmath}
\usepackage{amssymb}
\usepackage{mathtools}
\usepackage{amsthm}
\usepackage{array}
\usepackage{arydshln}

\usepackage[capitalize,noabbrev]{cleveref}

\theoremstyle{plain}

\theoremstyle{definition}

\theoremstyle{remark}

\usepackage[textsize=tiny]{todonotes}

\usepackage[utf8]{inputenc} 
\usepackage[T1]{fontenc}    
\usepackage{hyperref}       
\usepackage{url}            
\usepackage{booktabs}       
\usepackage{amsfonts}       
\usepackage{nicefrac}       
\usepackage{microtype}      
\usepackage{xcolor}         
\usepackage{bbold}
\usepackage{bm}
\usepackage{graphicx}
\usepackage{enumitem} 

\def\0{\bm{0}} 
\def\1{\bm{1}}
\newcommand\diag[1]{{\text{diag}\left(#1\right)}} 

\def\1{\bm{1}}

\def\rvb{{\mathbf{b}}}

\def\rmH{{\mathbf{H}}}

\def\rmW{{\mathbf{W}}}
\def\rmX{{\mathbf{X}}}

\def\vmu{{\bm{\mu}}}
\def\vsigma{{\bm{\sigma}}}

\def\vb{{\bm{b}}}

\def\mH{{\bm{H}}}

\def\mW{{\bm{W}}}
\def\mX{{\bm{X}}}

\DeclareMathAlphabet{\mathsfit}{\encodingdefault}{\sfdefault}{m}{sl}
\SetMathAlphabet{\mathsfit}{bold}{\encodingdefault}{\sfdefault}{bx}{n}

\icmltitlerunning{HadaNorm: Diffusion Transformer Quantization through Mean-Centered Transformations}

\begin{document}

\twocolumn[
\icmltitle{HadaNorm:\\Diffusion Transformer Quantization through Mean-Centered Transformations}



\icmlsetsymbol{equal}{*}

\begin{icmlauthorlist}
\icmlauthor{Marco Federici}{equal,qualcomm}
\icmlauthor{Riccardo Del Chiaro}{equal,qualcomm}
\icmlauthor{Boris van Breugel}{qualcomm}
\icmlauthor{Paul Whatmough}{qualcomm}
\icmlauthor{Markus Nagel}{qualcomm}
\end{icmlauthorlist}

\icmlaffiliation{qualcomm}{Qualcomm AI Research}

\icmlcorrespondingauthor{Marco Federici}{mfederic@qti.qualcomm.com}
\icmlcorrespondingauthor{Riccardo Del Chiaro}{rdelchia@qti.qualcomm.com}

\icmlkeywords{Machine Learning, ICML}

\vskip 0.3in
]



\printAffiliationsAndNotice{\icmlEqualContribution} 

\begin{abstract}
Diffusion models represent the cutting edge in image generation, but their high memory and computational demands hinder deployment on resource-constrained devices. Post-Training Quantization (PTQ) offers a promising solution by reducing the bitwidth of matrix operations. However, standard PTQ methods struggle with outliers, and achieving higher compression often requires transforming model weights and activations before quantization.
In this work, we propose HadaNorm, a novel linear transformation that extends existing approaches by both normalizing channels activations and applying Hadamard transforms to effectively mitigate outliers and enable aggressive activation quantization.
We demonstrate that HadaNorm consistently reduces quantization error across the various components of transformer blocks, outperforming state-of-the-art methods.
\end{abstract}

\section{Introduction}
\comment{
\begin{enumerate}
    \item Introduce the goal: efficient inference with diffusion models
    \item Introduce quantization as a method
    \item Introduce the issues of naive quantization and the solutions in literature
    \item Mention the existing shortcomings and introduce our method.
    \item Enumerate the contributions
\end{enumerate}
}

Diffusion models have emerged as the leading technique in deep learning for image 
generation, offering unparalleled visual realism.
However, this advancement comes at a significant computational cost, primarily 
due to the large model sizes and the iterative denoising procedures required for 
each image generation.
As the demand for scalable and efficient deployment of these models grows both on
the cloud and on the edge where computational resources are scarce,
optimizing inference efficiency has become a critical area of research.

Post-Training Quantization (PTQ) presents a promising solution for enhancing inference
efficiency by quantizing weights and activations, especially for high-power demanding 
operations such as linear layers and large matrix multiplications.
Despite its potential, PTQ faces substantial challenges, particularly when pushing 
activation bitwidth to 8-bits (A8) or weight to 4-bits (W4).
Outliers are a major hurdle for low-bitwidth quantization: they can only be represented when a large quantization grid is used, leading to large bins and loss of precision for the majority of data. One approach to reduce outliers and consequently improve quantization is to apply invertible
transformations to weights and activations, which do not alter the overall model output, but which do allow scaling outlier channels \cite{Xiao23}. Follow-up works \cite{Ashkboos24, Liu24, Ma24,Zhao25} use fast Hadamard transforms to mix channels, which effectively whitens the distribution and reduces outliers.
\begin{figure}
    \centering
    \includegraphics[width=0.95\linewidth]{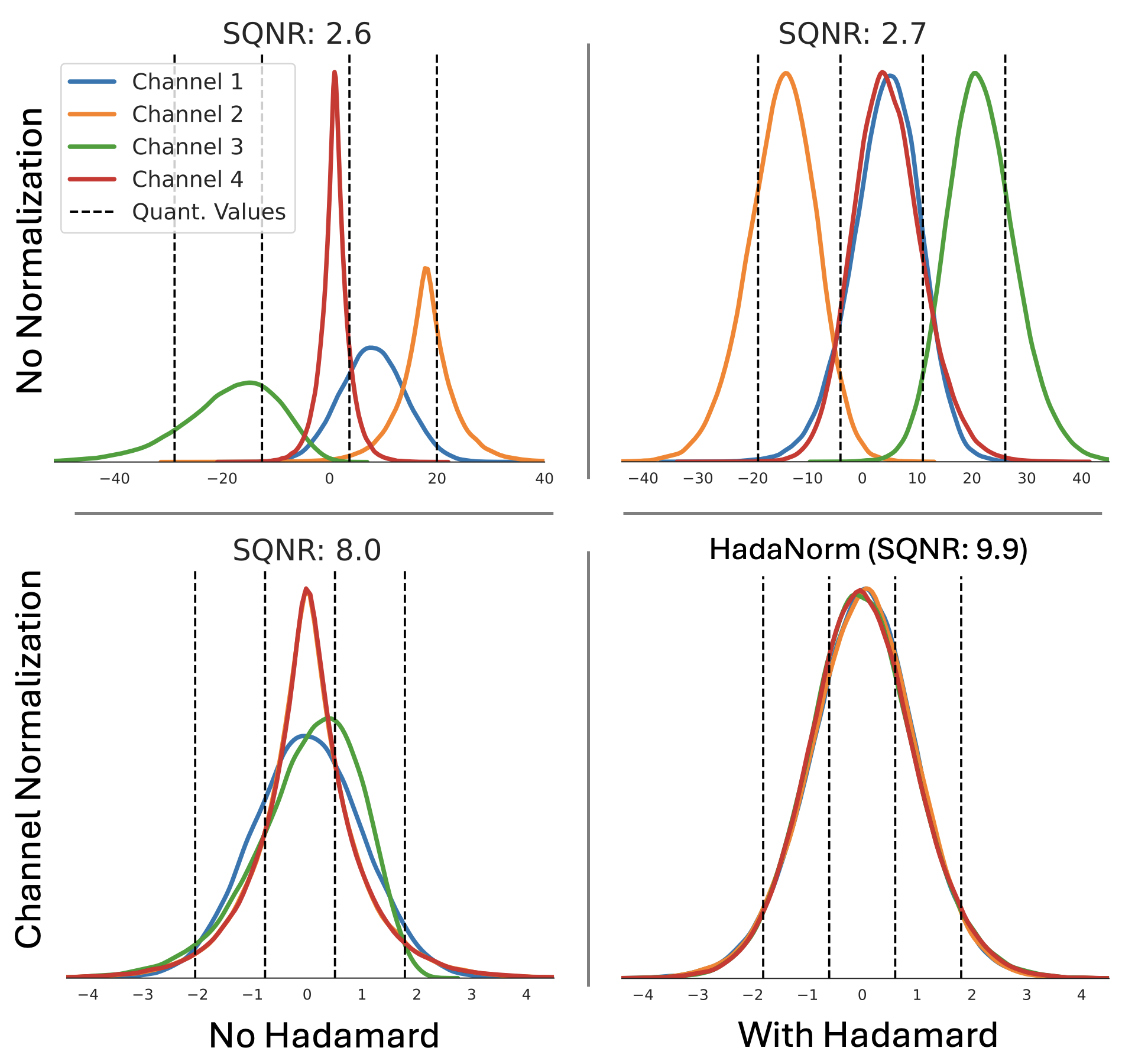}
    \caption{\textbf{HadaNorm reduces quantization error.} We take an illustrative setting of four channels with different distributions (top-left). Normalization (bottom-left) improves quantization, but it does not mix channels and hence cannot get rid of heavy tails. Hadamard transform (HT) (top-right) suffers when the channels have different means. HadaNorm (bottom-right) achieves better whitening, by both normalizing and applying the HT. The more Gaussian is easier to quantize.}
    \label{fig:main_idea}
\end{figure}

This paper extends previous work by introducing a simple yet effective centering transformation, which can be combined with existing approaches to further reduce quantization error.
%

\paragraph{Contributions.}
Our contributions are as follows:
\begin{enumerate}[itemsep=0pt,topsep=0pt]
    \item We argue why the efficacy of the Hadamard transform for activation quantization is reduced due to mean and scale differences across channels (Figure \ref{fig:main_idea}).
    \item We introduce HadaNorm, a simple transformation that applies centering and rescaling of channels together with a Hadamard transform to further improve performance of activation quantization. 
    \item Empirically, we show that HadaNorm significantly improves quantization performance in image diffusion transformers (DiT), resulting in a state-of-the-art CLIP score of 31.69 on PixArt-Sigma at W4A4.
\end{enumerate}

\section{Related Work}
Transformations for better quantization are most explored in the Large Language Models (LLM) space. \citet{Xiao23} find that activation quantization is harder than weight quantization, and propose per-channel scales to move outliers from activations to weights.

\citet{Ma24, Shao24} propose an affine transformation, that effectively adds a static bias term for LLM quantization. \citet{Ashkboos24} propose to use Hadamard transforms, which are cheap at inference time, yet allow spreading of outliers across channels. \citep{Zhao25} combines channel scaling and mixing to achieve higher compression on DiT architectures, although mixed precision is required to address quantization-sensitive components in the architecture.

\citet{Li24} further extends previous work on channel scaling \citet{Xiao23} by introducing high-precision low-rank matrices to absorb the weight outliers prior to weight quantization, achieving higher compression levels at the cost of a small overhead. \citet{Shao25} defines time-dependent transformation to address changes in the activation distribution during subsequent denoising steps, by applying transformation tailored to the activation distributions based on \citet{Lin240}, at the cost of additional overhead at inference time.
Our method further extends the literature on transformation applied to DiT models by introducing a simple centering strategy that does not require additional calibration and introduces minimal overhead, yet consistently improves upon existing strategies.

\comment{
    \begin{enumerate}
        \item ViDiT SmoothQuant+QuaRot \citep{Zhao25}
        \item TR-DQ: time-based rotation using DuQuant \citep{Shao25}
        \item SVDQuant: SVD + quantized residuals \citep{Li24}
    \end{enumerate}
}
\section{Method}
\subsection{Distributional differences between channels reduces effectiveness of per-token quantization}
Some of the related works use Hadamard transforms to mix channels, thereby spreading outliers over multiple channels and making each channel distribution approach a Gaussian distribution \citep{Liu2025}. This can be intuitively motivated by the central limit theorem: the Hadamard transform (HT) effectively adds up different channels (with signs flipped at times).

We make the simple observation that, after applying HT, channels may still exhibit substantially different mean. This is often the case in vision transformer models, as channels tend to have substantially different moments. Since all channels are quantized using the same quantization grid, this results in sub-optimal quantization.

We illustrate this using a toy example. In Figure \ref{fig:main_idea} top-left, we plot the distribution of four channels. If we naively choose a quantization grid based on all four channels, we see that the quantization error is large, as many channels collapse to just one or two values. When we apply a HT (top-right), we find that indeed each channel is closer to a normal distribution. However, due to non-zero mean of the initial channels, we observe large difference in the means across channels after applying the HT. For example, channel 2 corresponds to the row $[+1,-1,+1,-1]$ of the HT, which due to the large positive mean of the initial channel 2, and large negative mean of the initial channel 3, leads to a large negative mean after the HT.

The solution is to apply both channel normalization and  Hadamard transform. In the bottom-left figure, we show the effect of normalizing the channels. Although the SQNR is significantly better than before, channels are not mixed and outliers can heavily affect the scale of the quantization grid. Normalization alone does not affect the kurtosis of the channel distributions. We propose HadaNorm, which combines the dynamic channel normalization with the HT. Though simple, we get the best of both worlds: the HT mixes the channel distributions to reduce heavy tails, and because channel distributions are normalized, per-token quantization is more effective. Let us consider how this can be applied to a transformer model.

\begin{figure*}[!tbh]
    \centering
    \begin{minipage}{0.4\textwidth}
        \includegraphics[width=\textwidth,trim={1cm 0.4cm 10cm 0},clip]{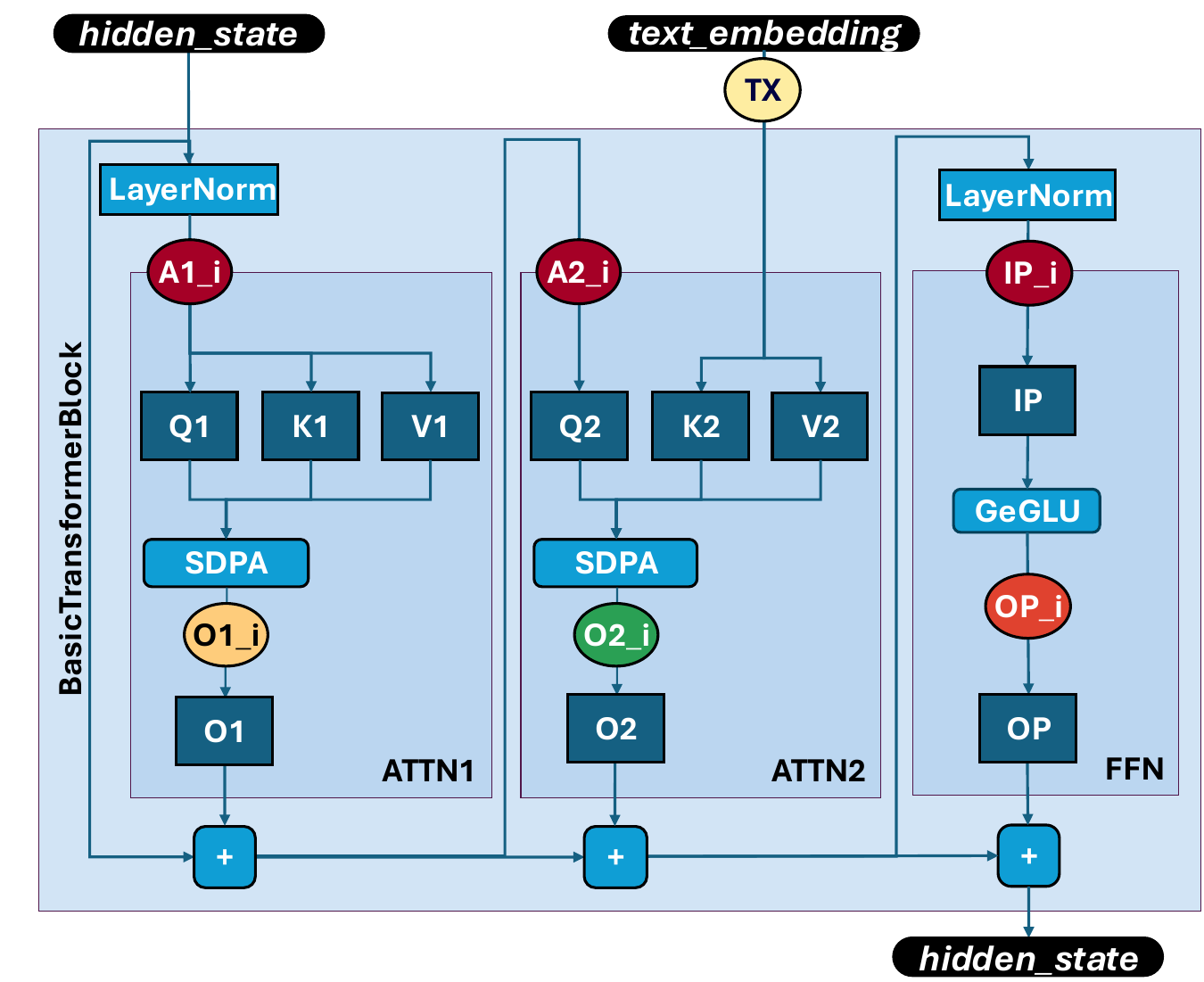}  
    \end{minipage}
    \begin{minipage}{0.4\textwidth}
        \includegraphics[width=\textwidth,trim={1cm 0.4cm 10cm 0},clip]{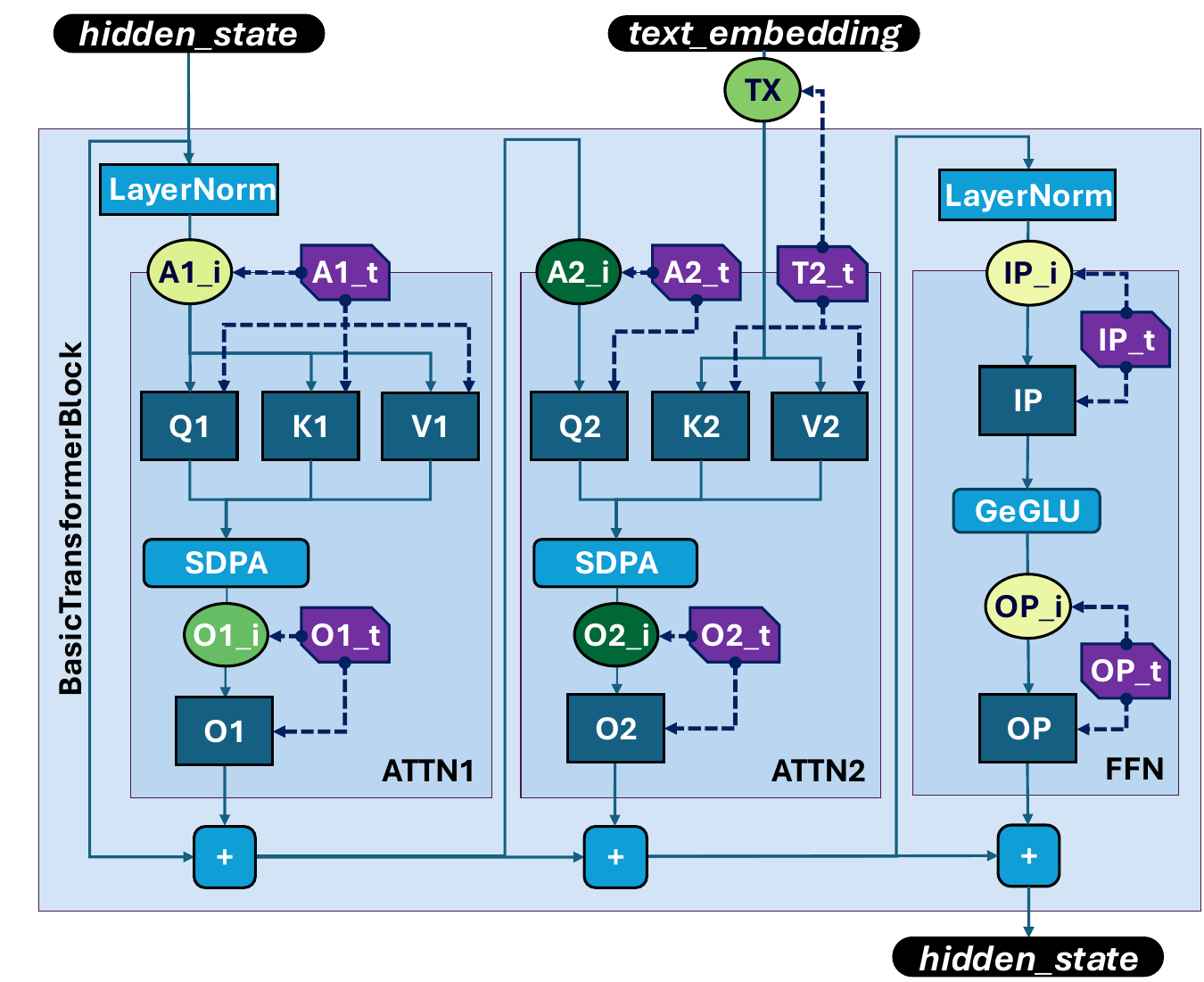} 
    \end{minipage}
    \begin{minipage}{0.08\textwidth}
        \includegraphics[width=0.65\textwidth, trim={15cm 1cm 0cm 0.5cm}, clip]{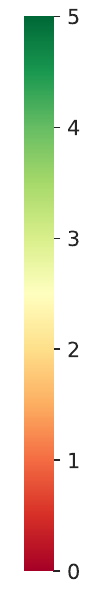}
    \end{minipage}
    \caption{\textbf{HadaNorm reduces quantization error of all quantizers.} Effect of activation quantization for various components of the DiT architecture without (left) and with (right) HadaNorm transformations (indicated in purple). Activation Quantizers (circles) are colored by the corresponding impact on the SQNR. The dynamic centering is not applied on the textual quantizer.}
    \label{fig:quant_error}
\end{figure*}

\subsection{HadaNorm}
To reduce the distributional differences between channels, ideally, we would want to both normalize the channels and apply the Hadamard transform. Let us consider what this means for a linear layer. Let us assume input vector $\rmX\in\mathbb{R}^{s\times d}$, a linear layer with weights $\rmW\in\mathbb{R}^{d\times m}$ and bias $\rvb\in\mathbb{R}^m$, a Hadamard transform represented as $\rmH\in\mathbb{R}^{d\times d}$, and assume vectors $\vmu,\vsigma\in\mathbb{R}^d$. We can write:
\begin{align}
    \rmX\rmW+\rvb = (\rmX\diag{\vsigma^{-1}}\mH)(\mH^T\diag{\vsigma}\rmW)+\rvb\nonumber\\
     =(\rmX\diag{\vsigma^{-1}}\mH-\vmu+\vmu)(\mH^T\diag{\vsigma}\rmW)+\rvb\nonumber\\
     =\underbrace{(\rmX\diag{\vsigma^{-1}}\rmH-\vmu)}_{\tilde{\rmX}}\underbrace{(\rmH^T\diag{\vsigma}\rmW)}_{\tilde{\rmW}}+\underbrace{(\rvb+\vmu\tilde\rmW)}_{\tilde{\rvb}}.
     \label{eq:hadanorm}
\end{align}
Thus, we can maintain the original model output, but both $\tilde{\rmX}$ and $\tilde\rmW$ have less outliers and will be quantized.

The per-channel mean $\vmu$ can be computed dynamically for each (transformed) batch element:
\begin{align}
    \vmu = \frac{1}{s}\1^T(\rmX\diag{\vsigma^{-1}}\mH).
    \label{eq:bias}
\end{align}
Note that the bias correction $\vmu\tilde\rmW$ can be computed efficiently in parallel, by appending a new token $\vmu$ to the transformed sequence $\tilde\mX$ before performing the matrix multiplication with $\tilde\mW$.

The scale $\vsigma$ cannot be chosen dynamically: this would require dynamically quantizing $\tilde{\rmW}$ depending on the current batch, which is undesirable. Instead, we follow \citet{Xiao23}, and statically determine $\vsigma$ based on the relative scales of $\rmX$'s channels and $\rmW$ input channels determined using a small calibration set:
\begin{align}
    \boldsymbol{\sigma}_i = \max(|\mX_i|)^{\alpha}/\max(|\mW_i|)^{1-\alpha}.
\end{align}

We place the HadaNorm layer throughout the network, before each linear layer (see Figure \ref{fig:quant_error}).

\section{Experiments}

Following \citet{Zhao25, Shao25, Li24}, we evaluate a quantized PixArt-Sigma \citep{Chen24} architecture on subset of captions from the COCO 2024 dataset \citep{Lin14} using 20 denoising steps. A disjoint calibration set is used to determine the activation statistics, which are used to tune the hyper-parameters $\alpha$.

\paragraph{Quantization} We quantize activations preceding each linear layer in the transformer blocks, which are indicated with ovals in Figure~\ref{fig:quant_error}, at 4 bits precision (A4). Following \citet{Zhao25}, we dynamically compute a separate quantization grid for each token based on the minimum and maximum values. Weights in linear layers are also quantized at 4 bits in blocks of size 128.

\paragraph{Metrics} We evaluate model performance using Signal to Quantized Noise Ratio (SQNR) computed in the latent space, the CLIP score \citep{Hessel21}, and CLIP IQA \citep{Wang23} as a proxy of visual image quality based on CLIP features \citep{Radford2021}.

\paragraph{Baselines}
We compare HadaNorm against other quantization transformation strategies proposed in recent litteraure including SmoothQuant \citep{Xiao23}, QuaRot \citep{Ashkboos24}, and the Static-Dynamic Channel Balancing (SDCB) method proposed in \citep{Zhao25}, which combines HT with channel-wise scaling.


\begin{figure*}[]
    \centering
    \newcommand{\fixedlabel}[1]{\vspace{-3.5mm}\parbox[c][4em][c]{\linewidth}{\centering #1}\vspace{-3mm}}
    \begin{minipage}{0.15\linewidth}
        \centering
        \fixedlabel{Original}\\
        \includegraphics[width=\linewidth]{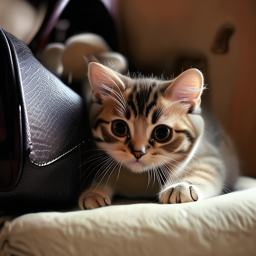}
    \end{minipage}
    \begin{minipage}{0.15\linewidth}
        \centering
        \fixedlabel{W4A4}\\
        \includegraphics[width=\linewidth]{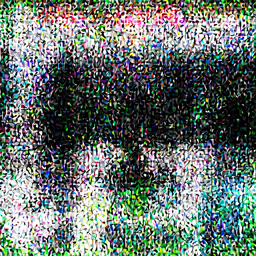}
    \end{minipage}
    \begin{minipage}{0.15\linewidth}
        \centering
        \fixedlabel{W4A4 + \\SmoothQuant}\\
        \includegraphics[width=\linewidth]{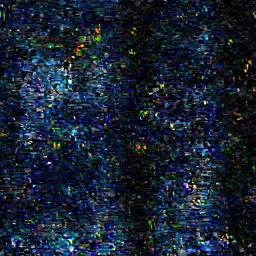}
    \end{minipage}
    \begin{minipage}{0.15\linewidth}
        \centering
        \fixedlabel{W4A4 + \\ QuaRot}\\
        \includegraphics[width=\linewidth]{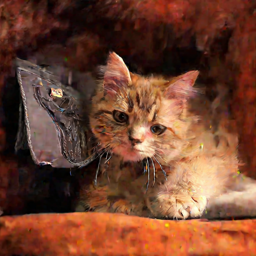}
    \end{minipage}
    \begin{minipage}{0.15\linewidth}
        \centering
        \fixedlabel{W4A4 +\\ SDCB}\\
        \includegraphics[width=\linewidth]{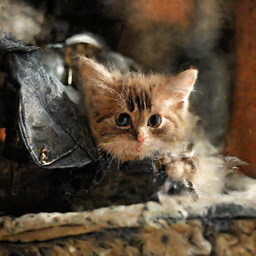}
    \end{minipage}
    \begin{minipage}{0.15\linewidth}
        \centering
        \fixedlabel{W4A4 + \\HadaNorm}\\
        \includegraphics[width=\linewidth]{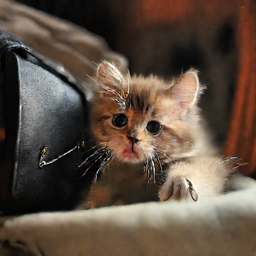}
    \end{minipage}
    \caption{Visualization of the denoised images for the W4A4 quantized model starting from the same noise input and the prompt "An adorable cat attempts to hide in a purse to steal the persons identity".
    \vspace{-1mm}
    }
    \label{fig:denoised_w4a4}
\end{figure*}

\subsection{HadaNorm reduces outliers throughout the network}

\paragraph{Set-up} 
First we aim to isolate the source of quantization error, and see how HadaNorm may help. In all transformer blocks we switch off all quantizers except one activation quantizer (at 4 bit), measure the quantization error in terms of SQNR at the end of the denoising process, and repeat this for all quantizers. Subsequently, we repeat this experiment, but with each transformation applied before each quantizer.

\begin{figure}
    \centering
    \includegraphics[width=1\linewidth]{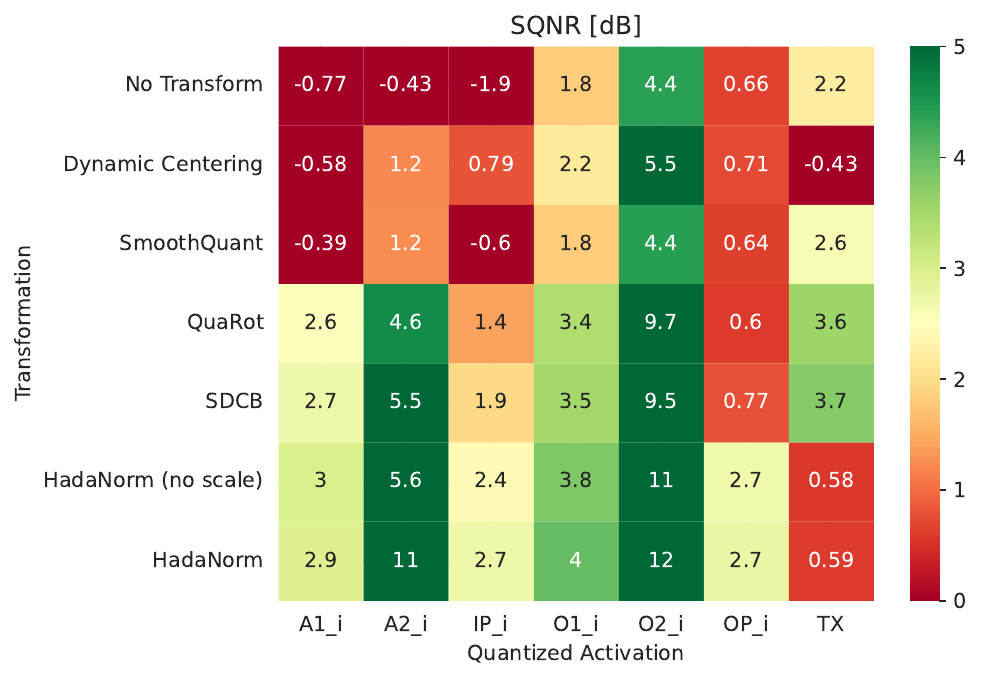}
    \vspace{-7mm}
    \caption{\textbf{HadaNorm's gain is mostly due to the combination of centering and the Hadamard transform.} Visualization of the SQNR resulting from the quantization of specific activations in the DiT architecture using various transformations.}
    \label{fig:single_act}
\end{figure}

\paragraph{Results.} We see (Figure \ref{fig:quant_error}, left) that the largest quantization error originates from the quantization of image inputs to the attention and FFN blocks, and the \texttt{OP} layer. Adding HadaNorm (right) consistently improves SQNR with the exception of the Text quantizer (\texttt{TX}). Figure \ref{fig:single_act} reports the result of the same analysis for a range of transformations. We find that the source of improvements can not be explained by one individual transform, but rather by their composition. In particular, this study shows that centering significantly improves the performance of HT.

\subsection{HadaNorm is SOTA at aggressive quantization}
\paragraph{Set-up.}
In this experiment we compare the end-to-end performance of HadaNorm when quantizing Pixart-Sigma to W4A4. Transforms are applied as indicated in Figure~\ref{fig:single_act} (right): each forward transform is applied online, while inverse channel scaling and HT are fused into the linear weights. Biases $\tilde\vb$ are also computed dynamically following the expression in Equation~\ref{eq:hadanorm} and Equation~\ref{eq:bias}.

\begin{table}[!tbh]
    \centering
    \setlength{\tabcolsep}{1pt}
        \caption{\textbf{HadaNorm outperforms all baselines.} Measure of SQNR [dB], CLIP score and CLIP IQA Pixart-Sigma models with 4 bits weights and activation quantization on a subset of the COCO 2024 dataset.\\}
    \label{tab:end2end}
    \begin{tabular}{lccc}
\toprule
\textbf{Transform} & \textbf{SQNR} $(\uparrow)$ & \textbf{CLIP score}$(\uparrow)$ & \textbf{CLIP IQA}$(\uparrow)$ \\
\midrule
Original & $\infty$ & 31.66 & 0.90 \\
No Transform & -2.88 & 19.22 & 0.11 \\\midrule
SmoothQuant & -2.03 & 18.79 & 0.12 \\
QuaRot & -0.39 & 30.88 & 0.76 \\
SDCB  & 0.01 & 31.17 & 0.84 \\
Dyn. Center & -2.32 & 19.68 & 0.14 \\
\textbf{HadaNorm} & 0.92 & 31.69 & 0.86 \\
\bottomrule
\end{tabular}
\end{table}

\paragraph{Results.}
We observe (Table \ref{tab:end2end})  that quantization is hard: without any transforms, the negative SQNR indicates that the noise exceeds the signal. SmoothQuant and dynamic channel centering alone do not help much as they cannot reduce outliers by spreading them over multiple channels. HTs (QuaRot) results in a significant improvement by mitigating the effect of outliers, although SQNR is still poor. HadaNorm gives significant further improvements over SDCB (HT + channel scaling) thanks to the additional dynamic centering operation.
 The visual outputs corresponding to the results reported in the table are visualized in Figure~\ref{fig:denoised_w4a4}.
 Additional results for 6-bits activation and weight quantization are reported in Appendix~\ref{app:additional_results}.


\section{Conclusion}
Although Hadamard transforms and channel scaling have been successfully used for improving quantization performance, we have shown that they are more effective when paired with a dynamic centering operation. The \textit{HadaNorm} transform is a promising tool for more aggressive quantization, whilst being simple to implement and cheap to use during inference. 
Although this work has focused on the transformer blocks in diffusion models, future work may explore whether the HadaNorm transform provides the same benefit for quantizing other models (e.g. LLMs), and non-transformer layers (e.g. CNN blocks).

\clearpage
\bibliographystyle{templates/icml/icml2025}
\bibliography{bibliography}

\newpage
\appendix
\onecolumn
\section{Additional results}
\label{app:additional_results}

\begin{table}[!tbh]
    \centering
    \setlength{\tabcolsep}{1pt}
        \caption{Measure of SQNR, CLIP score and CLIP IQA Pixart-Sigma models with 6 bits weights and activation quantization on a subset of the COCO 2024 dataset.}
    \label{tab:w6a6_results}
    \begin{tabular}{lccc}
\toprule
\textbf{Transform} & \textbf{SQNR} $(\uparrow)$ & \textbf{CLIP score}$(\uparrow)$ & \textbf{CLIP IQA}$(\uparrow)$ \\
\midrule
Original & $\infty$ & 31.66 & 0.90 \\
No Transform & 0.5 & 32.39 & 0.92 \\\midrule
SmoothQuant & 1.46 & 31.95 & 0.91 \\
QuaRot & 2.32 & 31.85 & 0.91 \\
SDCB & 2.61 & 31.81 & 0.91 \\
Dyn. Center & 1.56 & 31.87 & 0.91 \\
\bf{HadaNorm} & 3.05 & 31.82 & 0.91 \\
\bottomrule
\end{tabular}
\end{table}

\begin{figure}[!h]
    \newcommand{\fixedlabel}[1]{\parbox[c][4em][c]{\linewidth}{\centering #1}\vspace{-3mm}}
    \centering
    \setlength{\tabcolsep}{2pt}
    \begin{tabular}{>{\centering\arraybackslash}p{0.15\linewidth}>{\centering\arraybackslash}p{0.15\linewidth}
                >{\centering\arraybackslash}p{0.15\linewidth}
                >{\centering\arraybackslash}p{0.15\linewidth}
                >{\centering\arraybackslash}p{0.15\linewidth}
                >{\centering\arraybackslash}p{0.15\linewidth}}
\fixedlabel{Original} & 
\fixedlabel{W4A4} & 
\fixedlabel{W4A4 +\\ SmoothQuant} & 
\fixedlabel{W4A4 +\\ QuaRot} & 
\fixedlabel{W4A4 +\\ SDCB} & 
\fixedlabel{W4A4 +\\ HadaNorm} \\
\includegraphics[width=\linewidth]{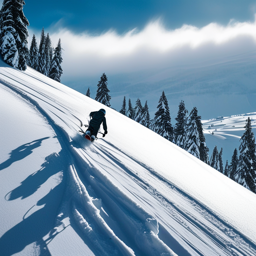} & 
\includegraphics[width=\linewidth]{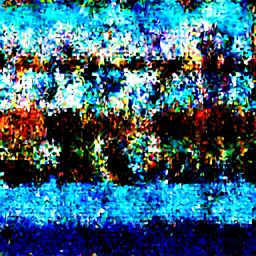} & 
\includegraphics[width=\linewidth]{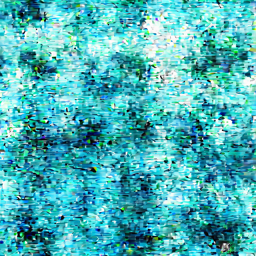} & 
\includegraphics[width=\linewidth]{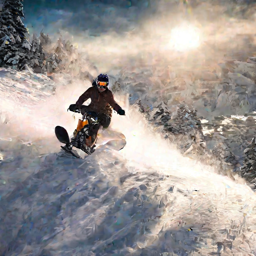} & 
\includegraphics[width=\linewidth]{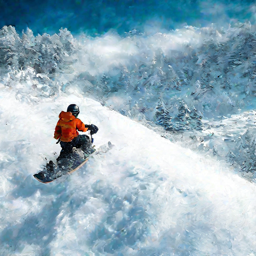} & 
\includegraphics[width=\linewidth]{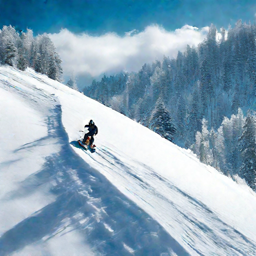}\\
\includegraphics[width=\linewidth]{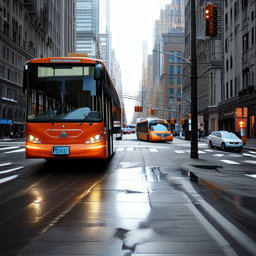} & 
\includegraphics[width=\linewidth]{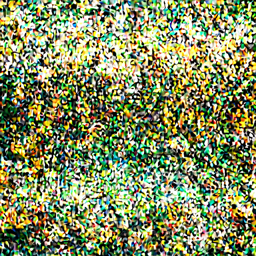} & 
\includegraphics[width=\linewidth]{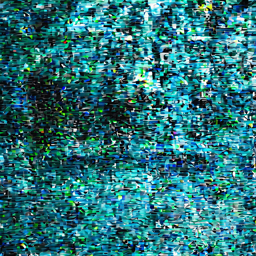} & 
\includegraphics[width=\linewidth]{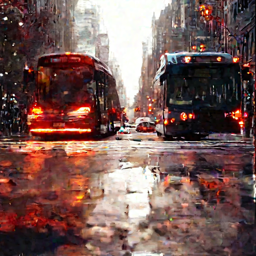} & 
\includegraphics[width=\linewidth]{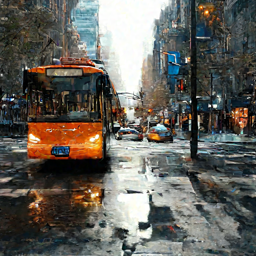} & 
\includegraphics[width=\linewidth]{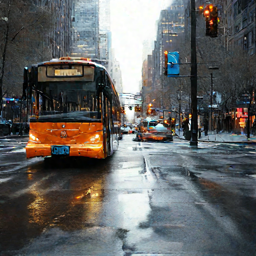}\\
\includegraphics[width=\linewidth]{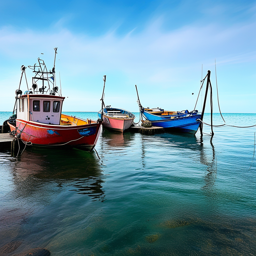} & 
\includegraphics[width=\linewidth]{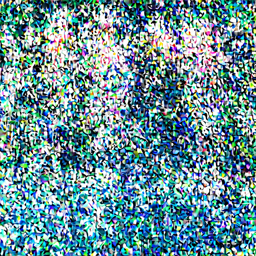} & 
\includegraphics[width=\linewidth]{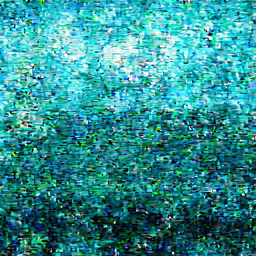} & 
\includegraphics[width=\linewidth]{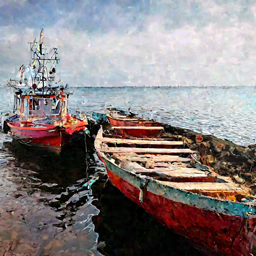} & 
\includegraphics[width=\linewidth]{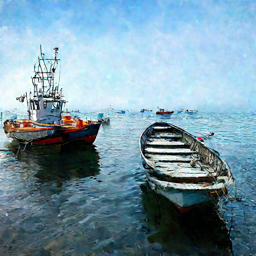} & 
\includegraphics[width=\linewidth]{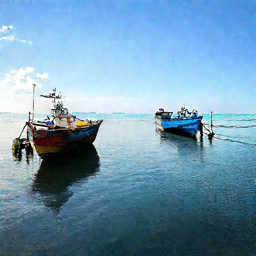}\\
\includegraphics[width=\linewidth]{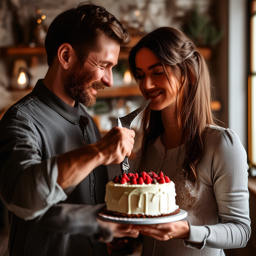} & 
\includegraphics[width=\linewidth]{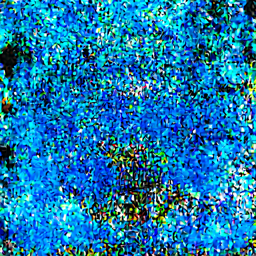} & 
\includegraphics[width=\linewidth]{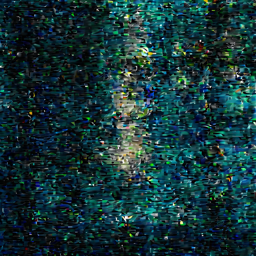} & 
\includegraphics[width=\linewidth]{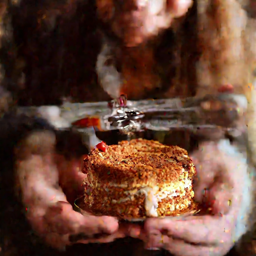} & 
\includegraphics[width=\linewidth]{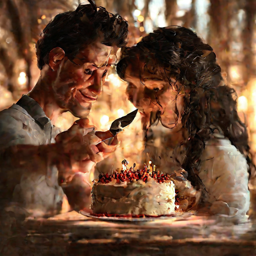} & 
\includegraphics[width=\linewidth]{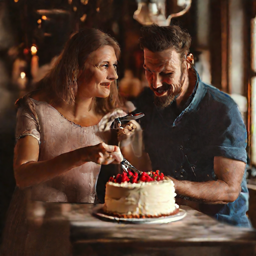}\\
\end{tabular}
    \centering
    \caption{Additional mage generations for the W4A4 Pixart-Sigma model with several transforms and COCO prompts.}
    \label{fig:additional_denoised_w4a4}
\end{figure}

\begin{figure}[!h]
    \newcommand{\fixedlabel}[1]{\parbox[c][4em][c]{\linewidth}{\centering #1}\vspace{-3mm}}
    \centering
    \setlength{\tabcolsep}{2pt}
    \begin{tabular}{>{\centering\arraybackslash}p{0.15\linewidth}>{\centering\arraybackslash}p{0.15\linewidth}
                >{\centering\arraybackslash}p{0.15\linewidth}
                >{\centering\arraybackslash}p{0.15\linewidth}
                >{\centering\arraybackslash}p{0.15\linewidth}
                >{\centering\arraybackslash}p{0.15\linewidth}}
\fixedlabel{Original} & 
\fixedlabel{W6A6} & 
\fixedlabel{W6A6 +\\ SmoothQuant} & 
\fixedlabel{W6A6 +\\ QuaRot} & 
\fixedlabel{W6A6 +\\ SDCB} & 
\fixedlabel{W6A6 +\\ HadaNorm} \\
\includegraphics[width=\linewidth]{figures/no/148272/no_transform.png} & 
\includegraphics[width=\linewidth]{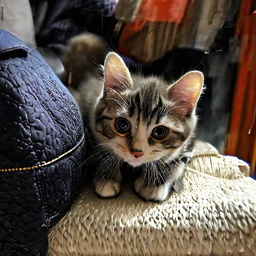} & 
\includegraphics[width=\linewidth]{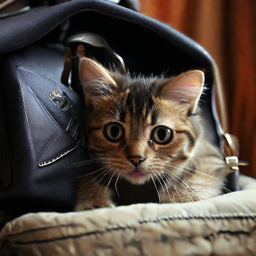} & 
\includegraphics[width=\linewidth]{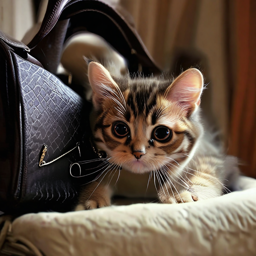} & 
\includegraphics[width=\linewidth]{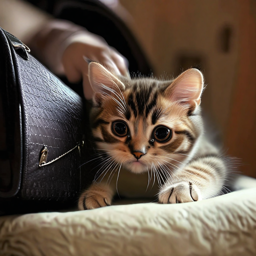} & 
\includegraphics[width=\linewidth]{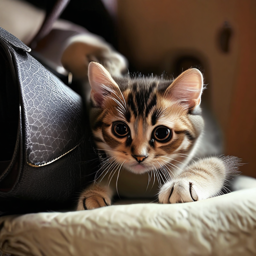} \\
\includegraphics[width=\linewidth]{figures/no/09171/no_transform.png} & 
\includegraphics[width=\linewidth]{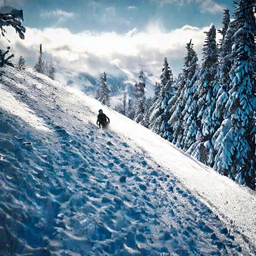} & 
\includegraphics[width=\linewidth]{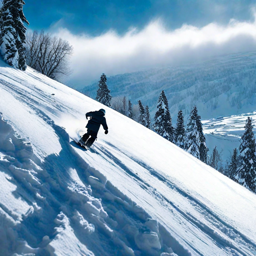} & 
\includegraphics[width=\linewidth]{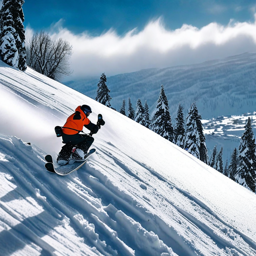} & 
\includegraphics[width=\linewidth]{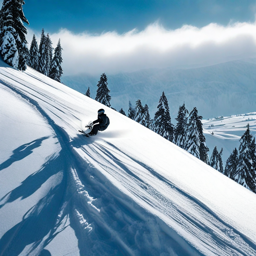} & 
\includegraphics[width=\linewidth]{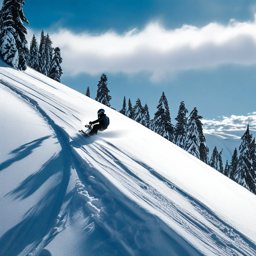}\\
\includegraphics[width=\linewidth]{figures/no/139549/no_transform.png} & 
\includegraphics[width=\linewidth]{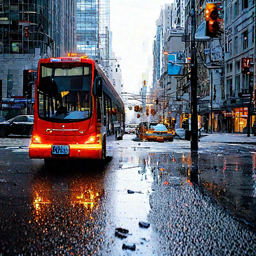} & 
\includegraphics[width=\linewidth]{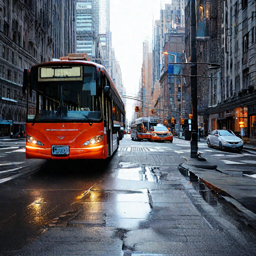} & 
\includegraphics[width=\linewidth]{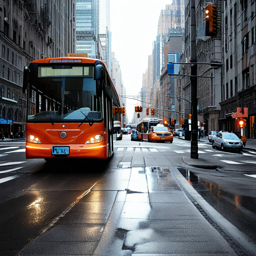} & 
\includegraphics[width=\linewidth]{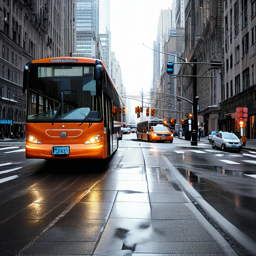} & 
\includegraphics[width=\linewidth]{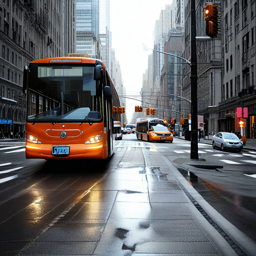}\\
\includegraphics[width=\linewidth]{figures/no/241758/no_transform.png} & 
\includegraphics[width=\linewidth]{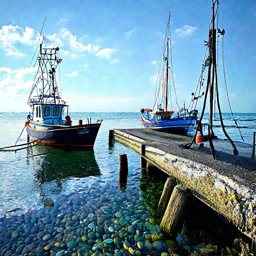} & 
\includegraphics[width=\linewidth]{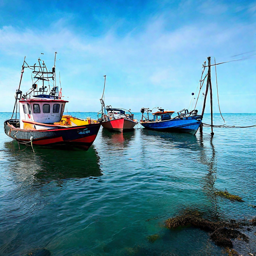} & 
\includegraphics[width=\linewidth]{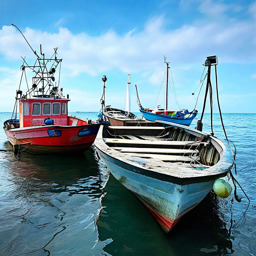} & 
\includegraphics[width=\linewidth]{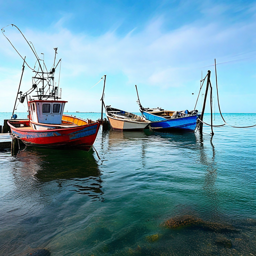} & 
\includegraphics[width=\linewidth]{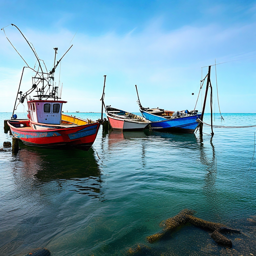}\\
\includegraphics[width=\linewidth]{figures/no/263969/no_transform.png} & 
\includegraphics[width=\linewidth]{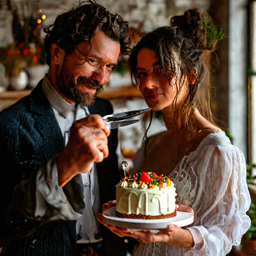} & 
\includegraphics[width=\linewidth]{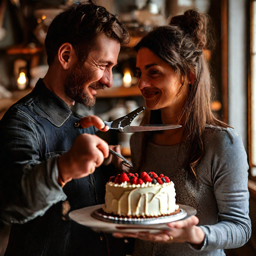} & 
\includegraphics[width=\linewidth]{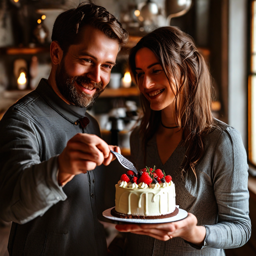} & 
\includegraphics[width=\linewidth]{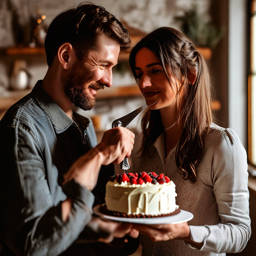} & 
\includegraphics[width=\linewidth]{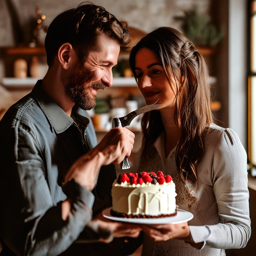}\\
\end{tabular}
    \centering
    \caption{Image generations for the W6A6 Pixart-Sigma model with several transforms and COCO prompts.}
    \label{fig:denoised_w6a6}
\end{figure}


\end{document}